\pdfoutput=1
\documentclass[runningheads]{llncs}
\usepackage[T1]{fontenc}
\usepackage{graphicx}
\usepackage{booktabs}
\usepackage[misc]{ifsym}
\newcommand{\corr}{(\Letter)}

\usepackage{amsmath,amssymb}
\usepackage{enumitem}
\usepackage{microtype}
\usepackage{hyperref}
\usepackage{multirow}

\usepackage[separate-uncertainty=true]{siunitx}
\usepackage{booktabs,makecell,multirow}
\renewrobustcmd{\boldmath}{}
\newrobustcmd{\B}{\fontseries{b}\selectfont}

\newcommand{\D}{\mathbf{D}}
\newcommand{\x}{\mathbf{x}}
\newcommand{\X}{\mathcal{X}}
\newcommand{\Y}{\mathcal{Y}}

\newcommand{\probability}{\mathbb{P}}
\newcommand{\parameters}{\mathbf{\Theta}}
\newcommand{\w}{\mathbf{w}}

\begin{document}

\title{Federated Learning of AnDE Classifiers}

\titlerunning{Federated Learning of AnDE Classifiers}

\author{
    Pablo Torrijos\inst{1,2}\orcidID{0000-0002-8395-3848} \corr \and
    Juan C. Alfaro\inst{1,2}\orcidID{0000-0003-1777-8540} \and
    Jos\'e A. G\'amez\inst{1,2}\orcidID{0000-0003-1188-1117} \and
    Jos\'e M. Puerta\inst{1,2}\orcidID{0000-0002-9164-5191}
}

\authorrunning{P. Torrijos et al.}

\institute{Departamento de Sistemas Inform\'aticos. Universidad de Castilla-La Mancha, 02071 Albacete, Spain.
    \and Instituto de Investigaci\'on en Inform\'atica de Albacete. Universidad de Castilla-La Mancha, 02071 Albacete, Spain.\\
\email{\{Pablo.Torrijos,JuanCarlos.Alfaro,Jose.Gamez,Jose.Puerta\}@uclm.es}}

\maketitle

\begin{abstract}
This work presents a federated framework for training Averaged $n$-Dependence Estimators (AnDE) in distributed environments. The proposed method focuses on the discriminative setting, where model weights are learned locally and aggregated globally, supporting any dependency order $n$. This design allows federated training without transmitting semantically meaningful parameters, improving privacy. Additionally, generative AnDE models are federated to provide a comparative baseline, with optional differential privacy applied to the aggregation of probability tables. Experiments on 12 discrete datasets show that discriminative models with $n \geq 1$ consistently outperform federated Naive Bayes (NB, $n=0$), and that privacy-preserving aggregation is effective with limited accuracy loss. These results establish federated AnDE as a viable and privacy-preserving framework, showing that probabilistic models remain applicable in modern federated learning settings.

\end{abstract}

%
%
\section{Introduction}\label{sec:introduction}

Federated Learning (FL)~\cite{mcmahan17aFL} enables collaborative model training across distributed clients without sharing raw data. While most FL research focuses on deep neural networks, structured probabilistic models offer complementary approaches with explicit control over dependency assumptions and parameter complexity. Naive Bayes (NB)~\cite{Webb2010} is a widely used probabilistic classifier based on strong independence assumptions that give simplicity and efficiency. The Averaged $n$-Dependence Estimator (AnDE)~\cite{Webb2011} generalizes NB by systematically relaxing this constraint, averaging over models where each attribute may depend on the class and up to $n$ other attributes. AnDE recovers  NB for $n=0$ and Averaged One-Dependence Estimators (AODE)~\cite{Webb2005} for $n=1$.

Existing FL approaches for NB typically adopt a generative formulation, aggregating local conditional probability tables with privacy mechanisms such as differential privacy~\cite{Giaretta2023}. However, these statistics are interpretable and may expose sensitive information. A safer alternative is to federate discriminative models that share only parameter weights, which carry less semantic meaning. NB$^w$ \cite{Zaidi2013}, a weighted discriminative version of NB, has recently been federated for $n=0$ \cite{Torrijos2024}, achieving strong privacy and accuracy. In this work, we investigate whether the strong independence assumption of NB can be relaxed in a federated setting.

To the best of our knowledge, no existing work addresses the federated training of AnDE classifiers. This paper fills that gap by introducing \textit{FedAnDE}, a federated learning framework for discriminative AnDE classifiers. FedAnDE extends NB$^w$ to models with a higher number of dependencies, allowing clients to compute and share only the weights. This improves model expressiveness while preserving data privacy. Additionally, we implement a federated version of the generative AnDE model, used as a baseline to assess the impact of discriminative training and differential privacy. Experimental evaluation shows that (i) increasing $n$ improves accuracy in federated discriminative models and (ii) the generative baseline suffers higher degradation under privacy constraints and as the number of clients increases.

%
%
\section{Background}\label{sec:background}

\paragraph{Averaged $n$-Dependence Estimators (AnDE)} \cite{Webb2011}.
Let the input space be $\X=\X_1\times\dots\times\X_d$, the class variable space $\Y=\{y_1,\dots,y_o\}$, and the training set $\D=\{(\x^{(i)},y^{(i)})\}_{i=1}^m$ with $\x=(x_1,\dots,x_d)\in\X$. Let $\mathcal{P}_n$ denote the set of all subsets $P\subseteq\{1,\dots,d\}$ of size $n$ with $0 \le n < d$. For each $P\in\mathcal{P}_n$, a $n$-dependence model factorizes the joint distribution for a given $y\in\Y$ as 
\begin{equation*}
    \probability(y,\x\mid P) =
    \probability(y)\,
    \probability(\x^{\downarrow P}\mid y)
    \prod_{j\notin P}\!\probability(x_j\mid y,\x^{\downarrow P}),
\end{equation*}
%
where $\x^{\downarrow P}$ denotes the projection of $\x$ onto $P$. This structure is known as a SuperParent-$n$-Dependence Estimator (SP$n$DE)~\cite{Webb2011}, where non-parent attributes are conditioned on the class and parent set $P$. Each $P$ defines a base classifier via the conditional distribution $\probability(y \mid \x, P)$, obtained by normalizing $\probability(y, \x \mid P)$ over $\Y$. AnDE combines these classifiers as a uniform ensemble:
\begin{equation*}
    \probability(y\mid\x) =  
    \frac{1}{|\mathcal{P}_n|}
    \sum_{P\in\mathcal{P}_n}
    \frac{\probability(y,\x\mid P)}{\sum_{y'\in\Y}\probability(y',\x\mid P)}.
\end{equation*}
For $n=0$ and $n=1$, the model reduces to NB and AODE, respectively. Larger $n$ captures more dependencies but increases complexity combinatorially.

\paragraph{Weighted Naive Bayes.}
NB$^w$~\cite{Zaidi2013} augments NB by attaching a weight $w_\theta$ to every generative parameter $\theta\in\parameters$. After estimating $\parameters$ generatively, the weight vector $\w$ is learned discriminatively by maximizing the conditional log-likelihood (CLL). Being $\parameters_{y,\x}$ the parameters used to score $(y,\x)$, the class label score becomes
\begin{equation*}
    \log\probability(y\mid\x) \propto \!\!
    \sum_{\theta\in\parameters_{y,\x}}\!w_\theta\,\log\theta.
\end{equation*}
\paragraph{Extending weights to AnDE.}
We extend NB$^w$ to AnDE$^w$ by assigning a separate weight set $\w_P$ to each $P \in \mathcal{P}_n$. For each $P$, the weighted $n$-dependence model is
\begin{equation*}
    \probability(y\mid\x,P,\w_P) \propto
    \probability(y)^{w_{0,y}}\,
    \probability(\x^{\downarrow P}\mid y)^{w_{P,y}}
    \prod_{j\notin P}\!
    \probability(x_j\mid y,\x^{\downarrow P})^{w_{j,y,\x^{\downarrow P}}}.
\end{equation*}
The ensemble prediction is the average $\probability(y \mid \x) = \frac{1}{|\mathcal{P}n|} \sum_{P \in \mathcal{P}_n} \probability(y \mid \x, P, \w_P)$.

\paragraph{Federated training.}
In a federated scenario, a generative AnDE model requires clients to transmit class-conditional counts, which, if not appropriately protected, are directly interpretable and may leak private information. With the weighted formulation, each client keeps its counts locally, optimizes the corresponding $\w_P$, and shares only these weights \cite{Torrijos2024}. Because $\w_P$ lacks semantic meaning, this strategy preserves privacy without needing extra protection steps, yet still lets the central server aggregate and update the global model.

%
%
\section{FedAnDE: Federated Weighted AnDE}\label{sec:fedande}

FedAnDE is a federated learning algorithm for discriminatively trained AnDE models. Each client $c$ holds a local training dataset $\D_c$ and trains its weighted model using only $\D_c$. The key idea is to keep all generative parameters $\parameters_c$ local and fixed and to share only discriminative weights $\w_{P,c}$ associated with each structure $P \in \mathcal{P}_n$. Let $C$ be the number of clients. For a given $P \in \mathcal{P}_n$, the server initializes the global weights $\w_P^{(0)}$, which are refined collaboratively over $T$ federated rounds. At each round:

\begin{enumerate}
    \item The server sends the current global weights $\w_P^{(t-1)}$ to all clients.
    \item Each client $c$ replaces local weights $\w_{P,c}$ with $\w_P^{(t-1)}$ and runs $L$ steps of \mbox{L-BFGS-B} optimizer \cite{Zhu1997} on $\D_c$ to maximize CLL, keeping $\parameters$ fixed.
    \item The updated local weights $\w_{P,c}^{(t)}$ are sent back to the server.
    \item The server aggregates the weights as:
    \begin{equation*}
        \w_P^{(t)} = \frac{1}{C} \sum_{c=1}^C \w_{P,c}^{(t)}.
    \end{equation*}
\end{enumerate}

This procedure is performed independently for each structure $P \in \mathcal{P}_n$, allowing for parallelization. The final prediction for a test instance $\x$ is obtained as:
\begin{equation*}
    \probability(y \mid \x) = \frac{1}{|\mathcal{P}_n|} \sum_{P \in \mathcal{P}_n} \probability(y \mid \x, P, \w_P^{(T)}).
\end{equation*}
Since only weights are communicated and the generative parameters remain private, FedAnDE improves privacy over standard federated generative AnDE, which relies on sharing class-conditional statistics, sensitive data that can still be reconstructed or inferred despite protective measures.

%
%
\section{Experimental Evaluation}\label{sec:experiments}
We evaluate FedAnDE with $n \in \{0,1,2\}$ on 12 categorical classification datasets from OpenML \cite{OpenML2013}. Their main properties are listed in Table~\ref{tab:datasets}. Each dataset is partitioned into $C \in \{5,10,20,50,100\}$ clients; larger $C$ values reduce local data per client, increasing sparsity and making the federated setting more challenging. All experiments use 5-fold cross-validation and are repeated five times with independent random seeds. We compare our proposed discriminative method, FedAnDE, with two baselines: (i) a federated generative variant with differentially private aggregation and (ii) client-local generative and discriminative models trained independently without collaboration.

\begin{table}[htbp]
\caption{Datasets used in the experimental evaluation. Here, $m$ is the number of instances, $d$ the number of attributes, and $o$ the number of classes.}\label{tab:datasets}
\resizebox{\textwidth}{!} {%
    \begin{tabular*}{0.6\textwidth}{@{\extracolsep{\fill}}lS[table-format=6.0]S[table-format=2.0]S[table-format=2.0]S[table-format=6.0]}
    \toprule
    \multicolumn{1}{c}{\multirow{2}{*}{\textsc{\bfseries Dataset}}} &\multicolumn{4}{c}{\textsc{\bfseries Properties}} \\
    \cmidrule(){2-5}
    & \multicolumn{1}{r}{\(m\)} & \multicolumn{1}{r}{\(d\)} & \multicolumn{1}{r}{\(o\)} & \multicolumn{1}{r}{\textsc{OpenML ID}}\\
    \midrule
        \textsc{House Votes 84} &         435 &           16 &         2    & \href{https://www.openml.org/search?type=data&id=56}{56} \\
        \textsc{Soybean} &         683 &           35 &        19   &  \href{https://www.openml.org/search?type=data&id=42}{42} \\
        \textsc{Tic-Tac-Toe} &         958 &            9 &         2   & \href{https://www.openml.org/search?type=data&id=50}{50}  \\
        \textsc{Flare} &        1066 &           11 &         6   &  \href{https://www.openml.org/search?type=data&id=46174}{46174} \\
        \textsc{Car Evaluation} &        1728 &            6 &         4    & \href{https://www.openml.org/search?type=data&id=991}{991} \\
        \textsc{Splice} &        3190 &           60 &         3    & \href{https://www.openml.org/search?type=data&id=46}{46} \\
    \bottomrule
    \end{tabular*}
    
    \hspace{0.4cm}
    
    \begin{tabular*}{0.6\textwidth}{@{\extracolsep{\fill}}lS[table-format=6.0]S[table-format=2.0]S[table-format=2.0]S[table-format=6.0]}
    \toprule
    \multicolumn{1}{c}{\multirow{2}{*}{\textsc{\bfseries Dataset}}} &\multicolumn{4}{c}{\textsc{\bfseries Properties}} \\
    \cmidrule(){2-5}
    & \multicolumn{1}{r}{\(m\)} & \multicolumn{1}{r}{\(d\)} & \multicolumn{1}{r}{\(o\)} & \multicolumn{1}{r}{\textsc{OpenML ID}}\\
    \midrule
        \textsc{Kr-vs-Kp} &        3196 &           36 &         2    &  \href{https://www.openml.org/search?type=data&id=3}{3} \\
        \textsc{Mushroom} &        8124 &           22 &         2    &  \href{https://www.openml.org/search?type=data&id=24}{24} \\
        \textsc{Phishing Websites} &       11055 &           30 &         2    & \href{https://www.openml.org/search?type=data&id=4534}{4534} \\
        \textsc{Nursery} &       12960 &            8 &         5    &  \href{https://www.openml.org/search?type=data&id=1568}{1568} \\
        \textsc{Kr-vs-K} &       28056 &            6 &        18    & \href{https://www.openml.org/search?type=data&id=46173}{46173} \\
        \textsc{Connect-4} &       67557 &           42 &         3    &  \href{https://www.openml.org/search?type=data&id=40668}{40668} \\
    \bottomrule
    \end{tabular*}
}
\end{table}

\paragraph{Federated discriminative training (FedAnDE).}
Training proceeds for $T\!=\!10$ rounds. In each round, the server sends the current weights to each client $c$, running $L\!=\!5$ L-BFGS-B \cite{Zhu1997} steps locally (with fixed $\parameters_c$) and returning updated $\w_{P,c}$. Only weight vectors are transmitted, which lack direct semantic meaning and reduce privacy leakage; no differential privacy mechanism is applied.

\paragraph{Generative baseline.} 
Generative models are trained in a single round by aggregating class-conditional counts from all clients. To ensure $\varepsilon$-differential privacy~\cite{Dwork2013DP}, each local count is perturbed with Laplace noise $\text{Lap}(\Delta_{n,d}/\varepsilon)$ before aggregation. In an $n$-dependence model, each instance updates: (1) one class count for $\probability(y)$; and (2) for each $P\in\mathcal{P}_n$, one count for $\probability(\x^{\downarrow P} \mid y)$ plus $d-n$ conditional counts for $\probability(x_j \mid y, \x^{\downarrow P})$, being $j \notin P$. The $\ell_1$-sensitivity is therefore:
\begin{equation*}
    \Delta_{n,d} \;=\; 1 \;+\; |\mathcal{P}_n| \cdot (1+d-n).
\end{equation*}
where $|\mathcal{P}_n| = \binom{d}{n}$. Noisy counts are normalized to yield valid probability tables. We fix $\varepsilon = 1.0$ for all experiments.

\paragraph{Client-local baseline.}
Each client trains independently on its local data. Generative models use maximum-likelihood estimation; discriminative ones are optimized with L-BFGS-B~\cite{Zhu1997} without iteration limits. Accuracy is averaged across clients.

\paragraph{Reproducibility.}
Datasets are publicly available on OpenML (see ID on Table \ref{tab:datasets}). All code (including centralized and federated variants of generative and discriminative models) and data are on GitHub\footnote{\url{https://github.com/ptorrijos99/BayesFL}}. Generative models build on the NB implementation from Weka\footnote{\url{https://ml.cms.waikato.ac.nz/weka/}} library, and discriminative ones extend NB$^w$\footnote{\url{https://github.com/nayyarzaidi/EBNC}}.

\subsection{Results}
\begin{figure}[htb]
    \centering
    \includegraphics[width=1\linewidth]{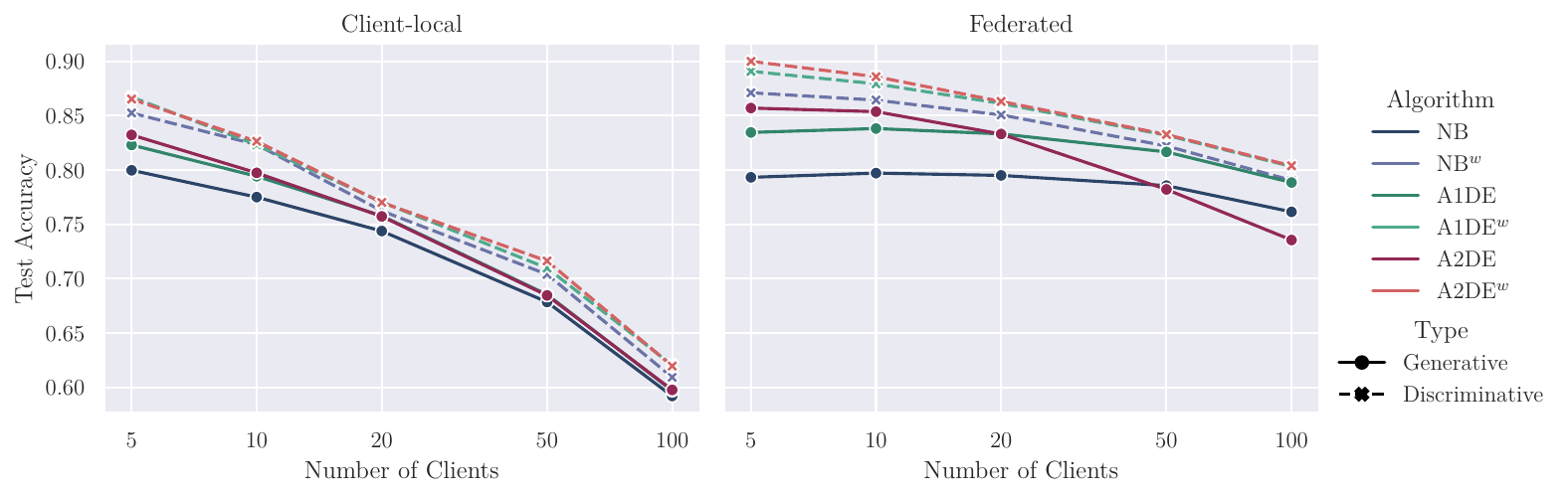}
    \caption{Mean test accuracy for generative (NB, A1DE, A2DE; solid lines) and discriminative (NB$^w$, A1DE$^w$, A2DE$^w$; dashed lines) models in client-local (left) and federated (right) settings with $C \in \{5,10,20,50,100\}$ clients.}
    \label{fig:acc-results}
\end{figure}

Figure~\ref{fig:acc-results} reports the mean test accuracy across datasets and the number of clients. Three trends are consistent across all scenarios. (i) Federated training significantly improves performance compared to client-local models, especially as the number of clients increases. Aggregating the clients' knowledge helps mitigate data scarcity at each client. (ii) Discriminative models consistently outperform generative ones, with NB$^w$, A1DE$^w$, and A2DE$^w$ dominating their unweighted counterparts (NB, A1DE, and A2DE). This supports the effectiveness of the CLL optimization of our FedAnDE framework. (iii) Increasing the dependency order improves accuracy. Moving from $n=0$ (NB) to $n=1$ (A1DE) and $n=2$ (A2DE) yields steady gains, which are more noticeable in federated learning since the higher the order, the more data tends to be used. This confirms that capturing attribute dependencies is beneficial even under federated constraints. 

Among all methods, A2DE$^w$ consistently achieves the highest accuracy across all federation levels. As the number of clients increases and local data per client decreases, performance naturally drops, and the gap between models narrows. In particular, generative approaches become more competitive in low-data regimes, reflecting the well-known tendency of generative models to outperform discriminative ones when data is scarce~\cite{Ng2001Discriminative}. Nonetheless, higher-order discriminative models remain robust and consistently lead to overall accuracy.

\begin{figure}[htb]
    \centering
    \includegraphics[width=1\linewidth]{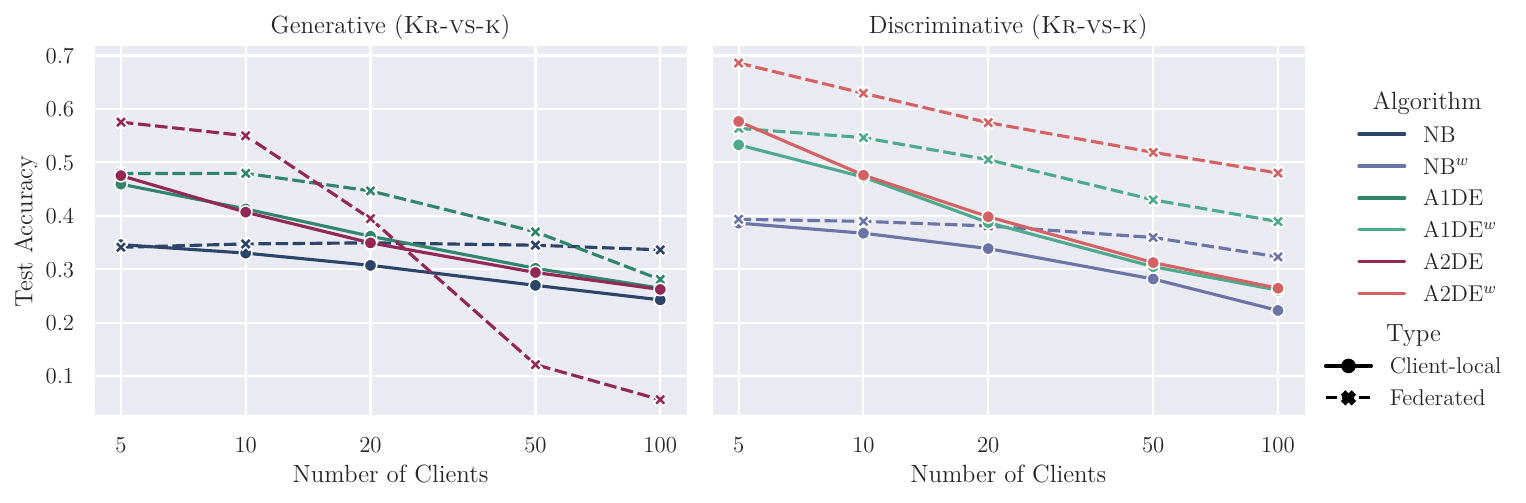}
    \caption{Test accuracy on \textsc{Kr-vs-K} for generative (NB, A1DE, A2DE; left) and discriminative (NB$^w$, A1DE$^w$, A2DE$^w$; right) models with $C \in \{5,10,20,50,100\}$ clients.}
    \label{fig:acc-results-King}
\end{figure}

\paragraph{Case Study: \textsc{Kr-vs-K} Dataset.} Figure~\ref{fig:acc-results-King} presents detailed results on the \textsc{Kr-vs-K} dataset, which is known for strong dependencies between attributes, such as piece positions in a chess endgame. This structure favors higher-order models: both in the generative (left) and discriminative (right) panels, A2DE consistently outperforms A1DE, which in turn outperforms NB. This confirms that increasing $n$ yields better performance when attribute interactions are key.

Federated training provides large improvements over client-local baselines across all values of $n$, and the gains are larger for more complex models. However, the generative variants show a clear drop as the number of clients increases. In particular, A1DE eventually performs worse than NB, and A2DE suffers a sharp drop from 20 clients. This is due to a combination of limited local data per client and the impact of differential privacy, which introduces more noise in higher-order models because they access more statistics. In contrast, the discriminative models remain more robust, with A2DE$^w$ again achieving the best results. Per-dataset results for all methods are reported in Appendix~\ref{app:results}.

%
%
\section{Conclusions and Future Work}\label{sec:conclusions}
We presented FedAnDE, a federated framework for training discriminative AnDE classifiers. By extending NB$^w$ to higher-order models and enabling weight-based aggregation, FedAnDE improves both accuracy and privacy. Experiments show that increasing $n$ improves performance, and that discriminative models outperform their generative counterparts.

While this work focuses on Bayesian classifiers, future comparisons with standard discriminative models such as logistic regression would help contextualize performance. While $L=5$ local steps and $T=10$ rounds provided stable results, a more detailed analysis of convergence, communication cost, and runtime is ongoing. Although $n=2$ yielded the best accuracy, its combinatorial cost may limit scalability for large $d$; future work will explore approximations such as sparse structures or dimensionality reduction. Additional directions include combining generative and discriminative updates, supporting continuous attributes through Gaussian AnDE, and evaluating alternative privacy mechanisms.

\begin{credits}
\sloppy
\subsubsection{\ackname} This work is partially funded by the following projects: TED2021-131291B-I00 (MICIU/AEI/10.13039/501100011033 and European Union NextGenerationEU/PRTR), SBPLY/21/180225/000062 (Junta de Comunidades de Castilla-La Mancha and ERDF A way of making Europe), PID2022-139293NB-C32 (MICIU/AEI/10.13039/501100011033 and ERDF, EU), FPU21/01074 \mbox{(MICIU/AEI/10.13039/501100011033} and ESF+); 2025-GRIN-38476 (Universidad de Castilla-La Mancha and ERDF A way of making Europe).

This version of the contribution has been accepted for publication, after peer review, but is not the Version of Record and does not reflect post-acceptance improvements, or any corrections. The Version of Record is published in: Koprinska, I., Mendes-Moreira, J., Branco, P. (eds) Machine Learning and Principles and Practice of Knowledge Discovery in Databases. ECML PKDD 2025. Communications in Computer and Information Science, vol 2841, pp. 464-471. Springer, Cham (2026), and is available online at: \url{https://doi.org/10.1007/978-3-032-19102-1_28}.

\subsubsection{\discintname}
The authors have no competing interests to declare that are relevant to the content of this article.

\end{credits}
%
%
%
\bibliographystyle{splncs04}
\bibliography{biblio}

%
%
\appendix
\renewcommand{\theHsection}{appendix.\Alph{section}} 
\section{Complete Results of the Experimental Evaluation} \label{app:results}
\begin{figure}[htbp]
    \centering
    \includegraphics[width=0.95\linewidth]{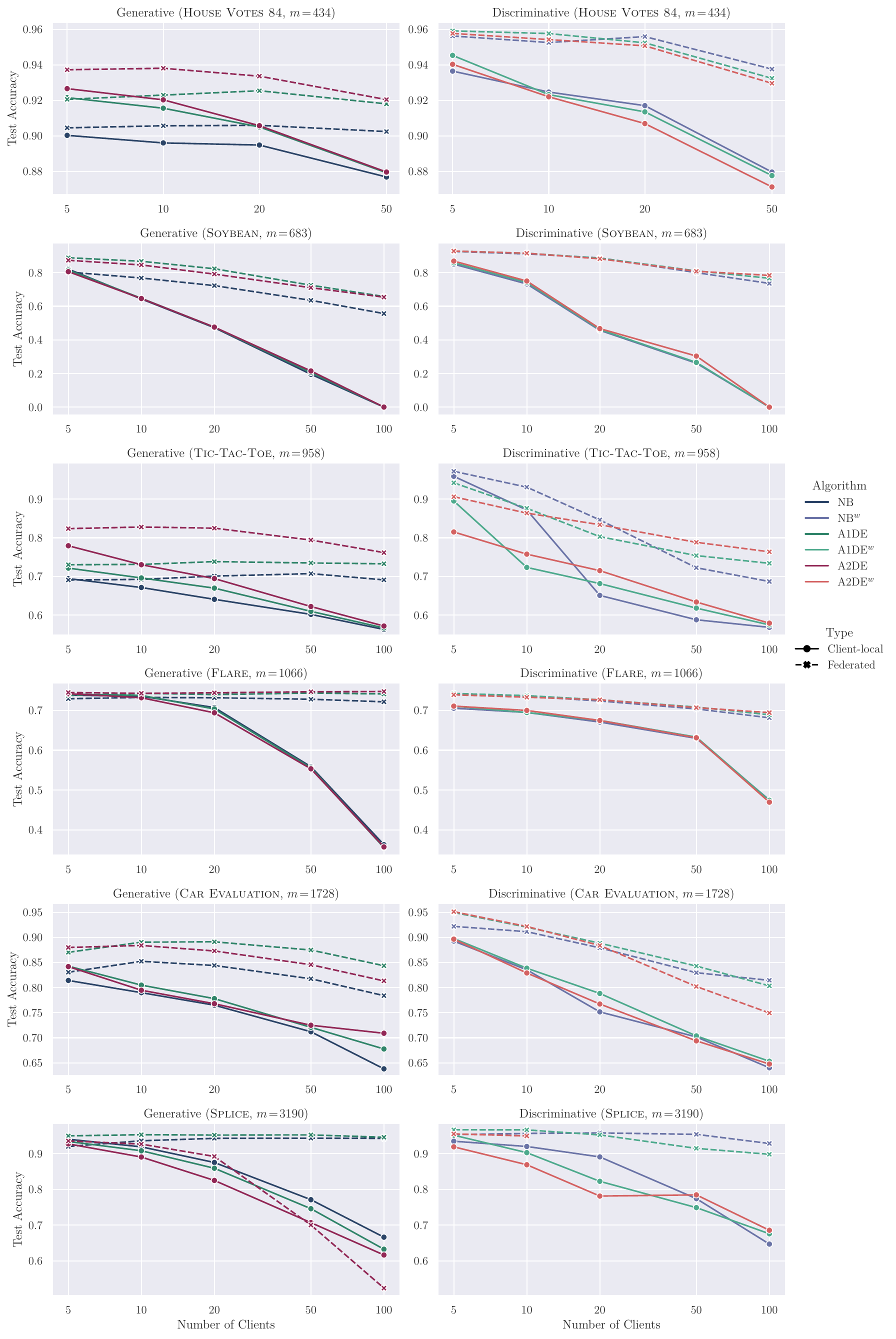}
    \caption{Accuracy results (part 1 of 2: first six datasets). Solid lines denote generative models (NB, A1DE, A2DE); dashed lines denote discriminative models (NB$^w$, A1DE$^w$, A2DE$^w$).}
    \label{fig:exp-first}
\end{figure}

\begin{figure}[htbp]
    \centering
    \includegraphics[width=0.95\linewidth]{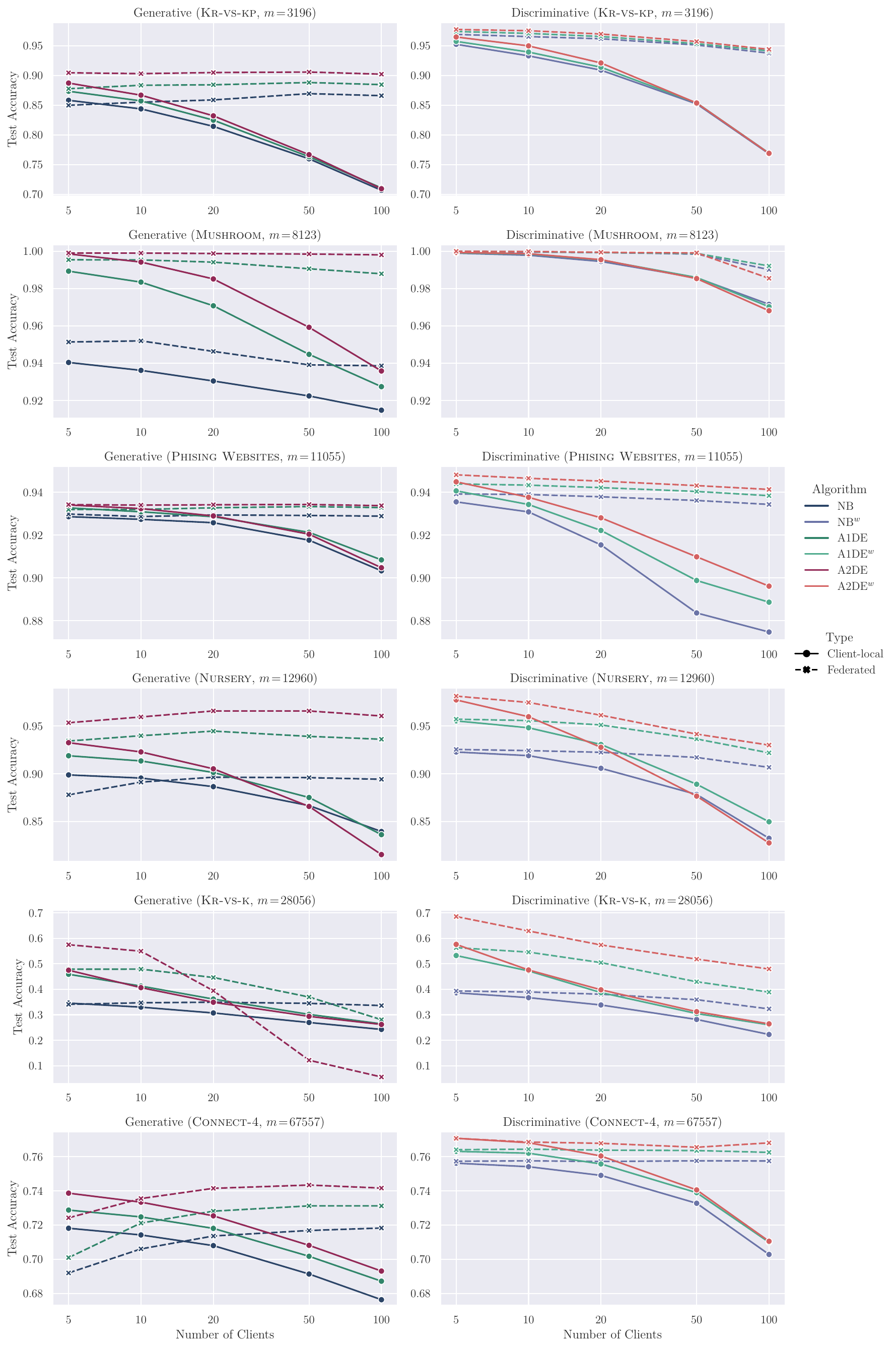}
    \caption{Accuracy results (part 2 of 2: last six datasets). Solid lines denote generative models (NB, A1DE, A2DE); dashed lines denote discriminative models (NB$^w$, A1DE$^w$, A2DE$^w$).}
    \label{fig:exp-last}
\end{figure}

Figures~\ref{fig:exp-first} and~\ref{fig:exp-last} report the accuracy obtained by every method on each dataset. The twelve datasets are split into two groups of six, ordered by the number of instances ($m$). The following trends are consistent, although with some dataset-specific nuances.

\paragraph{Federated versus local.}
All federated variants outperform their client-local counterparts. The performance gap widens as the number of clients increases, confirming the benefit of aggregating local updates under data fragmentation.

\paragraph{Discriminative versus generative.}
FedAnDE ($n \in \{0,1,2\}$) consistently matches or surpasses generative models. The advantage becomes more stable on larger datasets, while smaller datasets exhibit higher variability.

\paragraph{Effect of dependency order and data size.}
The expected trend A2DE~$>$~A1DE~$>$~NB is generally observed, particularly on large datasets such as \textsc{Nursery}, \textsc{Kr-vs-K}, and \textsc{Connect-4}. On small datasets like \textsc{Flare} or \textsc{Soybean}, the differences between $n$ values are negligible. In cases like \textsc{House Votes 84}, increasing $n$ improves accuracy in the generative models, but has a limited or inconsistent effect in the discriminative ones. This behaviour is expected: higher-order models require sufficient data to estimate their probability tables reliably, and their performance degrades when data is too sparse, especially under strong federation ($C = 100$).

\paragraph{Privacy impact.}
Generative models degrade due to the Laplace noise added to class-conditional counts, especially for higher-order models where each instance updates a larger number of statistics. However, the effect of the number of clients $C$ is non-monotonic. When $C$ is small, each client has enough data to produce reliable local estimates, but the aggregation averages over a few noisy vectors, so the residual noise remains large. Conversely, when $C$ is large, the averaging attenuates the noise (variance scales as $1/C$), but each client holds fewer samples, degrading the quality of the raw statistics even before noise is added. The result is a non-convex trade-off in accuracy, often peaking at intermediate values of $C$. This pattern is especially pronounced in datasets like \textsc{Connect-4}.

These results reinforce the main conclusion: higher-order discriminative AnDE models can be trained effectively in federated settings. Their performance scales well with data size and number of clients, offering stronger privacy guarantees than generative approaches.

\end{document}